\documentclass{article}
\usepackage{iclr2016_conference,times}
\usepackage{cite}
\usepackage{url}

\usepackage[cmex10]{amsmath}
\usepackage{bm}
\usepackage{textcomp}
\usepackage{eqparbox}
\usepackage{color}
\usepackage{floatflt}
\DeclareMathOperator*{\argmin}{arg\,min}
\usepackage{enumerate}
\usepackage{array}
\usepackage{multirow}
\usepackage{tabularx}
\newcolumntype{C}[1]{>{\centering}m{#1}}
\usepackage[caption=false,font=footnotesize]{subfig}
\usepackage{graphicx}
\usepackage{amsfonts}
\usepackage{amsmath}
\usepackage{amssymb,bm}

\title{$L_1$ logistic regression as a feature selection step for training stable classification trees for the prediction of severity criteria in imported malaria}

\author{Luca Talenti\\
\'{E}cole Centrale Paris\\
Chatenay-Malabry, 92290, France\\
\texttt{luca.talenti@student.ecp.fr} \\
\AND
Margaux Luck \& Anastasia Yartseva  \\
Institut HyperCube\\
Puteaux, 92907, France\\
\texttt{\{margaux.luck,anastasia.yartseva\}@institut-hypercube.org} \\
\And
Nicolas Argy \& Sandrine Houz\'{e} \\
Laboratoire de parasitologie, Hopital Bichat-Claude Bernard\\
Paris, 75018, France \\
\texttt{\{nicolas.argy,sandrine.houze\}@bch.aphp.fr} \\
\And
Cecilia Damon\\
Institut HyperCube\\
Puteaux, 92907, France\\
\texttt{cecilia.damon@institut-hypercube.org} \\
}

\begin{document}
\bibliographystyle{iclr2016_conference}

\maketitle

\begin{abstract}
Multivariate classification methods using explanatory and predictive models are necessary for characterizing subgroups of patients according to their risk profiles. Popular methods include logistic regression and classification trees with performances that vary according to the nature and the characteristics of the dataset. In the context of imported malaria, we aimed at classifying severity criteria based on a heterogeneous patient population. We investigated these approaches by implementing two different strategies: \emph{L1} logistic regression (\emph{L1LR}) that models a single global solution and classification trees that model multiple local solutions corresponding to discriminant subregions of the feature space. For each strategy, we built a standard model, and a sparser version of it. As an alternative to pruning, we explore a promising approach that first constrains the tree model with an \emph{L1LR}-based feature selection, an approach we called \emph{L1LR}-Tree. The objective is to decrease its vulnerability to small data variations by removing variables corresponding to unstable local phenomena. Our study is twofold: i) from a methodological perspective comparing the performances and the stability of the three previous methods, i.e \emph{L1LR}, classification trees and \emph{L1LR}-Tree, for the classification of severe forms of imported malaria, and ii) from an applied perspective improving the actual classification of severe forms of imported malaria by identifying more personalized profiles predictive of several clinical criteria based on variables dismissed for the clinical definition of the disease. The main methodological results show that the combined method \emph{L1LR}-Tree builds sparse and stable models that significantly predicts the different severity criteria and outperforms all the other methods in terms of accuracy. The study shows that new biological and epidemiological factors may be integrated in the current clinico-biological picture to improve diagnosis and patient treatment.
\end{abstract}

\section{Introduction}
For the purpose of diagnosis, the use of explanatory multivariate classification tools is essential to efficiently characterize groups of patients with a high risks of developing the disease \citep{Austin:2013}. Two off-the-shelf classifiers are linear logistic regression and decision trees. Based on two different model learning strategies, both methods investigate the relationship between binary response variables and a set of heterogeneous explanatory variables. To remove non-relevant variables and/or limit the complexity of the solution, both methods allow the integration of a penalization term in its objective function.
The choice of the $L_1$ penalization for logistic regression aims to reduce the risk of overfitting induced by potential co-linearity and the combinatorial exploration of all possible two-way interactions \citep{HosmerJr:2013,Park:2011}. $L_1$ logistic regression uses linear combinations of explanatory variables to learn a single decision boundary and build an easily understandable linear model. It selects a subset of discriminant features and assesses the predictive contribution of each of them in the model. However, it only considers linear interactions between features and their global variation related to the binary outcome. Moreover, it does not take into account missing data. 
The decision tree approach, non-parametric and non-linear,  is  particularly helpful to explore which feature subspaces are predictive of a class of subjects \citep{Breiman:1984, Atkinson:2000}. It learns multiple decision boundaries parallel to feature axes and builds easy-to-interpret models under the form of a set of if-then-else decision rules. It also handles data as complex as missing values, numerical and categorical data, multi-colinear variables, outliers and local relationships among variables. However, decision trees could generate over-complex and locally optimal solutions increasing the overfitting risk and ungystable decision trees due to small variations in the data. Pruning is generally applied to avoid the overfitting phenomenon.
Some studies have compared these two approaches showing that their relative performances and stability depend of the nature of the signal (e.g., the signal-to-noise ratio) and the characteristics of the dataset (size) \citep{Perlich:2003}. Therefore, recent studies have tried to combine them, particularly by applying a logistic regression model at the leaves of the decision trees in order to smooth the final model as an alternative to the standard pruning \citep{Landwehr:2005}. Another popular and efficient way to produce more accurate and stable classifiers is feature selection \citep{Guyon:2003}. In this study, we proposed and tested a novel approach, called $L_1$LR-Tree, that combines the two previous strategies. The objective of this novel approach is to fit more robust and simple decision tree learners by first applying a feature selection resulting from the $L_1$ logistic regression model. 

We carried out this methodological study in the particular context of a thorough understanding of the mechanisms of severe forms of imported malaria.
Despite the decrease of malaria cases in endemic areas since 2000 \citep{WHO}, the increasing number of travelers between endemic regions and western countries promotes imported malaria cases in non-endemic areas. Metropolitan France is the most concerned European country and the mortality rate of imported malaria is strongly related to severe malaria form favored by delays in access to health care. The World Health Organization (WHO) defined the different clinical and biological criteria for severe malaria in order to speed up the diagnosis and health care of patients that require urgent and intensive care units \citep{WHO}. This clinico-biological picture inferring the diagnosis is multi-criteria and complex \citep{SPILF, OMS}. It also does not take into account epidemiological information which could provide further insight. Indeed, contrary to endemic regions in Africa, the populations of patients with imported malaria is heterogeneous and composed of first generation migrants (born in endemic regions and living in France), second generation migrants (children of first generation migrants born and living in France) and African or European travelers, adults or children with a different history of malaria and genetic background. We also observed an epidemiological evolution in the clinical presentation of severe forms of imported malaria with an increase of older patients from a migrant background and a decrease in the number of patients having neurological disorders \citep{Seringe:2011, CNR}. 

Therefore, in this applied research work, we explored the influence of factors (demographic, epidemiological, clinical, biological and transcriptomic) dismissed in the current clinico-biological picture on both the diagnosis of severe forms of imported malaria and some clinical observations of acute malaria attacks (hematological syndrome, visceral failure, neurological disorders and  parasitaemia level). Risk factors for developing severe malaria have been investigated in this context of heterogeneous populations of patients with classical univariate statistical methods \citep{Seringe:2011}. However, these methods revealed their limits 
as they only assess the statistical associations between each factor and the severe criteria of imported malaria independently of each other and without prediction assumptions. Hence, the use of explanatory multivariate classification tools is essential to efficiently characterize groups with a high risk of developing complicated imported malaria and to reduce the mortality of \emph{Plasmodium falciparum} infection in France based on multi-source data \citep{Kajungu:2012}. Our comparative study of the different learning strategies is especially interesting when considering the complexity of the data. Indeed, the available datasets corresponding to the different case studies present several difficulties that need to be overcome: heterogeneous populations of patients with under-represented subgroups, local phenomena with a weak signal-to-noise ratio, missing data, bias in the current classification, small dataset, etc.  

In section~\ref{sec:MatMeth}, we first present the data and the 6 case studies grouped into two experiments, and then, we briefly explain:  the three classification methods, the model selection and the evaluation methodologies. In section~\ref{sec:Results}, we describe the different performance results and the learned models. Finally, in section~\ref{sec:Concl}, we conclude on the benefit of the combined method $L_1$LR-Tree, the clinical aspects and the perspectives of this work.

\section{Materials and Methods}
\label{sec:MatMeth}

\subsection{Dataset}
The French National Reference Center of Malaria (FNRCM\footnote{http://www.cnrpalu-france.org/}) monitors imported malaria for epidemiological purposes through a national network of correspondents in hospital centers. In a prospective manner, demographic (age, sex, ethnic origin, medical history, history of malaria, chemoprophylaxis taken), epidemiological (native country, country of residence, visited area), clinical (history of the disease, severity criteria, management of the patient, treatment), biological (severity criteria, biochemical parameters, hematological parameters, diagnostic tools, serological status) and transcriptomic (parasite genome) data have been collected in a secured database. The objective of this monitoring is to identify high-risk groups for the development of severe malaria.

\begin{table}[h!]
\caption{List of input variables by data type \label{tab:var}}
\begin{tabularx}{\textwidth}{ |X|X| }
\hline
\multicolumn{2}{|c|}{\textbf{Input Variables}} \\
\hline
\centering\textbf{Data type} & \centering{\textbf{Variables name \& description}} \\
\tabularnewline
\hline 
Demographic & Age, Sex, Caucasian (dichotomous), African (dichotomous), Chemoprophylaxis taken (dichotomous) \\
\hline
Epidemiological & Vis West Africa (Visit in West Africa, dichotomous), Vis Central Africa (Visit in Central Africa, dichotomous), Vis Other (Visit in an other endemic country, dichotomous), Res France or other non-endemic country (Resident in France or in another non-endemic country, dichotomous)\\
\hline
Clinical & ATCD (history of the disease), Delay 2 (days from $1^{st}$ symptoms to recovery), Immunodependency (dichotomous)\\
\hline
Biological & GB (White Blood Cells count), Platelets (platelets count), Serology, Serological Interpretation, Titration\\
\hline
Transcriptomic & A1, A2, A3, B1, B2, C1,C2, BC1, BC2, Var1, Var2csa, Var3 (parasite genome, i.e. expression of \textit{var} genes) \\
\hline
\end{tabularx}
\end{table}

We defined six case studies, called \textit{cs} in this paper, grouped into two experiments. The first experiment, composed of two case studies, explores the influence of demographic, epidemiological, clinical biological and transcriptomic factors dismissed in the current clinico-biological picture (see Table \ref{tab:var}) on the diagnosis of severe forms of imported malaria. The second experiment is composed of four case studies and explores the influence of the same previous factors on four clinical observations of acute malaria attacks: hematological syndrome, visceral failure, neurological disorders and  parasitaemia level. Note that for these two experiments, we have removed from the input variables those that are directly used to infer the target, which is the malaria severity degree, such as organ or metabolic dysfunctions and blood smear measures.

\paragraph{First experiment} The first case study focuses on the current diagnosis of severe imported malaria. For the second case study, we distinguished two subgroups of patients among those with severe malaria according to the existence of neurological and multi-organ clinical dysfunctions. The first one is called serious imported malaria and the second one is called critical imported malaria because this last form of the disease has a high probability of being fatal.

Finally, the studied dataset is composed of 353 patients diagnosed with three severity levels of imported malaria: moderate, serious or critical. For each patient, we have a total of 29 features. 12 of them concern the parasite's genome, giving information of different nature and sources. 

We define two case studies, each one comparing two groups of subjects: 
\begin{enumerate}[cs 1)]
\item 
the first case study includes the whole dataset by comparing subjects having moderate imported malaria to those having a severe form of the disease (i.e. serious and critical). 353 subjects are included in this experiment with 202 patients having a moderate form and 151 having a severe form. The objective of this experiment is to identify risk factors predictive of severe malaria in a heterogeneous population of patients.
\item 
the second case study compares subjects suffering from a serious imported malaria to those having a critical form of the disease. 151 subjects are included in this experiment with 88 having a serious form and 63 displaying the critical form. The objective of this experiment is to characterize and to validate the relevance of these two subgroups among patients suffering from severe malaria.
\end{enumerate}

\paragraph{Second experiment}The 4 different case studies consist in discriminating between two clinical states used for the definition of severe forms of imported malaria among a set of 343 patients. 

\begin{enumerate}[cs 1)]
\setcounter{enumi}{2}
\item 
the third case study compares patients suffering from Hematological Syndrome with people not affected by this condition. 49 patients suffer from this syndrome in the dataset.
\item 
the fourth case study compares patients suffering from Visceral Failure with people not affected by this condition. 271 patients suffer from this failure in the dataset.
\item 
the fifth case study compares patients suffering from Neurological Disorders with people not affected by this condition. 32 patients suffer from these disorders in the dataset.
\item 
the sixth case study compares patients displaying Parasitaemia greater than 4\% with people not affected by this condition. 113 patients display this condition in the dataset.
\end{enumerate}

\subsection{Methods}

\subsubsection{$L_1$ logistic regression}
For the classification step, we used an $L_1$ regularized logistic regression\footnote{Computed in R with the package glmnet, https://cran.r-project.org/web/packages/glmnet/glmnet.pdf} (i.e. $L_1LR$) modeling the class membership probability as a linear combination of explanatory variable \citep{HosmerJr:2013, Park:2011}. 
Standard logistic regression (i.e. non penalized) estimates a binary decision function by assuming that the logit can be modeled as a linear function of features: 
\begin{align}
&\log\left(\frac{p_{\beta}(\mathbf{x_{i}})}{1-p_{\beta}(\mathbf{x_{i}})}\right)=\eta_\beta(\mathbf{x_{i}}) \quad \text{with},\\
&\eta_\beta(\mathbf{x_{i}})=\beta_0 +\sum_{j=1}^{q}\mathbf{x}_{i,j}^{\top}\beta_{j},\nonumber\\ &p_{\beta}(\mathbf{x_{i}})=\mathbb{P}_\beta\left(Y =1|\mathbf{x_{i}}\right)\nonumber
\end{align}
$Y \in \{0,1\}$ is the binary target, $X=\{x_1,\cdots, x_q\} \in \mathbb{R}^{n}$ are the explicative variables, $\beta_0$ is the intercept and $\boldsymbol{\beta}\in \mathbb{R}^{q}$ is the regressor vector.

The $L_1$ penalization parameter is introduced in the model to shrink the estimates of the regression coefficients towards zero and set some of them to zero relative to the maximum likelihood estimates:
\begin{equation}
\boldsymbol{\hat\beta}=\argmin_{(\beta_{0},\boldsymbol{\beta})\in \mathbb{R}^{q+1}} F_{\lambda}(\beta_0,\boldsymbol{\beta})=-\l(\beta_{0},\boldsymbol{\beta})+\lambda\,\|\boldsymbol{\beta}\|_{1}
\end{equation}

where $l\left(\beta_0,\boldsymbol{\beta}\right)$ is the log-likelihood function:
\begin{equation}
l\left(\beta_0,\boldsymbol{\beta}\right)=\frac{1}{N}\sum_{i=1}^{N}y_{i}\left(\beta_{0}+\mathbf{x}_{i}^{\intercal}\boldsymbol{\beta}\right) - \log\left( 1 + e^{\beta_{0}+\mathbf{x}_{i}^{\top}\boldsymbol{\beta}}\right) 
\end{equation}

Note that the parasite genome data have not been included in $L_1LR$ method as it requires not-empty features and many subjects have missing values for these features. \\

\subsubsection{Decision Trees}
Decision tree analysis \footnote{Computed in \texttt{R} with the package \texttt{rpart}, \texttt{https://cran.r-project.org/web/packages/rpart/rpart.pdf}} successively splits the dataset into increasingly homogeneous subsets by binary recursive partitioning of multidimensional covariate space  until it is stratified to meet a splitting criterion \citep{Breiman:1984, Atkinson:2000}. The splitting criterion used is $SS_T - \left(SS_L +SS_R \right)$, where $SS_T = \sum_{i} \left(y_{i} - \bar y\right)^{2} $ is the sum of squares for the node and $SS_R, SS_L$ are the sums of squares for the right and left children respectively. This is equivalent to choosing a split that maximizes the between-groups sum of squares in an analysis of variance. We constrained a minimum number of 10 observations in the leaf nodes in order to avoid over-fitting.\\

\subsubsection{$L_1$LR-Tree}

The combined model consists of two steps:
\begin{itemize}
\item Select a subset of features by fitting a $L_1$ logistic regression. 
\item Build a decision tree on the selected features.
\end{itemize}
The purpose of this combined approach is the same as the pruning tree approach: to limit the appareance of an over-complex solution and lower the over-fitting risk. However, the penalization of these two approaches occurs in two different ways. Instead of pruning the learned decision tree to limit its size, the $L_1LR$-Tree approach prior constrains the model by reducing the dimension of input features.\\

\subsubsection{Model selection}
For both methods, we applied a model selection to limit the complexity of the solution with a $K$-fold cross-validation. We chose $K=\lfloor\frac{n}{10}\rfloor$ to both ensure a biais-variance trade-off of the test error estimates and a sufficient representation of the two groups within the test sets \citep{Kohavi:1995}. In each experimental dataset, we kept the original proportion of the two classes within each fold. 
For $L_1$ logistic regression, we optimized the penalization coefficient $\lambda$ so that we capture two levels of model complexity. 
Therefore, we selected two values of $\lambda$: $\lambda_{min}$ such that $\lambda_{min}$-$L_1LR$ is the best model minimizing the $K$-folds cross-validation mean squared error and $\lambda_{1se}$ such that $\lambda_{1se}$-$L_1LR$ corresponds to the simplest model which is no more than one standard error worse than the best model according to the one standard error rule \citep{Friedman:2001}.

For decision trees, we applied a cost-complexity minimization to limit the size of the tree and we called this simplified tree, the pruned tree $T_{\alpha}$:
\begin{align}
&T_{\alpha} = \argmin_T R_{\alpha}\left(T\right), \\
&R_{\alpha}\left(T\right)= R\left(T\right) + \alpha |T| \nonumber
\end{align}
The $\alpha\in [0,\infty[$ parameter defines the cost  of adding another split to the model. This parameter is optimized within the $K$-fold cross-validation.\\

\subsubsection{Evaluation}

As the output is dichotomous and the two regression methods, $L_1LR$ and decision trees, estimate the class membership probability $\hat y=\hat p(\mathbf{x_{i}})=\mathbb{P}\left(Y =1|\mathbf{x_{i}}\right)$, we define the following decision function to classify our samples:

\begin{equation}
\text{class}(\hat y) = 
\begin{cases}
    0, \quad \text{if} \quad \hat y \leq 0.4 \\
    1, \quad \text{if} \quad \hat y > 0.4
\end{cases}
\end{equation}

We fixed the threshold of the decision function with respect to the distribution of the two classes in the six case studies of the two experiments in order to avoid biased models due to imbalanced classes. For all the case studies "cs 1 to 6", the threshold is equal to the proportion of symptomatic patients. Hence, the patients are less frequently classified in the predominant class $0$ in a way that more strongly penalizes the misclassication of this class . 

We checked the predictive power of our constructed classifiers, based on the $L_1LR$, the regression trees and the $L_1LR$-tree methods, through three performance indicators: 

\begin{itemize}
\setlength\itemsep{1em}
\item \emph{Recall}:
\hphantom{\emph{Specificity}:\emph{Accuracy}:}
\scalebox{1.3}{ $\frac{\text{True Positives}}{\text{Positives}}$}

\item \emph{Specificity}:
\hphantom{\emph{Recall}:\emph{Accuracy}:}
\scalebox{1.3}{$\frac{\text{True Negatives}}{\text{Negatives}}$}

\item \emph{Accuracy}:
\hphantom{\emph{Recall}:\emph{Specificity}:}
\scalebox{1.3}{$\frac{\text{Recall}+\text{Specificity}}{2}$}
\end{itemize}

The \emph{Recall} (resp. \emph{Specificity}) score aims to quantify the overall rate of samples correctly classified for the second (resp. first) class. They give two complementary insights about the quality of classification performances of the different methods. Indeed, from a medical point of view, we aim to discriminate and well classify the two groups of patients and not only the predominant class. 
We also assessed the statistical significance of the recall and specificity scores with a binomial test and we defined three significance levels: * $p-value\leq 5\%$, ** $p-value\leq 1\%$,*** $p-value\leq 0.1\%$.

These performance indicators are computed with leave-one-out validation, classifying each patient one time, in order to generate stable learning models (highly correlated). This choice is due to the high heterogeneity of the patients.
 
\section{Results}
\label{sec:Results}

In a first part (\textit{First experiment}) we presented the results of our methodological objective, that is the comparative study of the different learning strategies, applied to the two case studies of the first experiment (moderate \textit{vs} severe malaria and serious \textit{vs} critical). We compared the results between the standard methods, i.e. $L_1$ LR and classification trees, and their sparse form, $\lambda_{min}$-$L_1LR$ \textit{vs} $\lambda_{1se}$-$L_1LR$ and Tree \textit{vs} Prune. Then, we selected the best forms of each standard methods in term of performance scores and sparsity and we combined them to build a $L_1$LR-Tree model.
In a second part (\textit{Second experiment}) we applied the combined model to the four case studies of the second experiment to validate this approach and point out some clinical insights given by the selected variables of the models. For all the case studies of the two experiments, we reported and explained the models obtained with the combined methods.

\subsection{First experiment}
\subsubsection{$L_1$ logistic regression- and regression trees-based models}
\paragraph{Performance scores}

\begin{figure}[h!]
\centering
\fbox{\includegraphics[width=3.5in]
{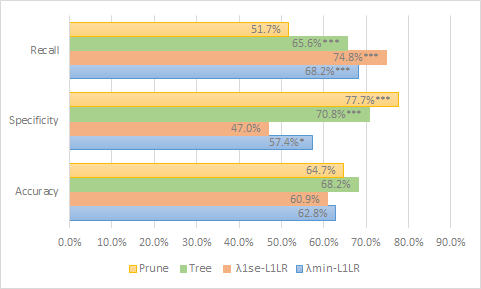}}
\caption{Moderate malaria form vs severe form ($1^{st}$ case study): comparison of the 4 models, Tree, Prune,  $\lambda_{min}$-$L_1LR$ and $\lambda_{1se}$-$L_1LR$, through the three leave-one-out performance scores: recall, specificity and accuracy. For each score-bar, the color refers to the model, the length indicates the score's value (rounded percent) with its significance level for the recall and specificity scores.}
\label{fig:stdperf4methods}
\end{figure}

\begin{figure}[h!]
\centering

\fbox{\includegraphics[width=3.5in]
{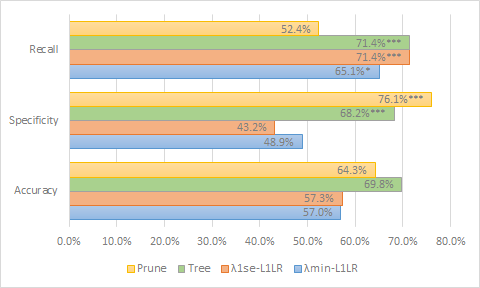}}
\caption{Serious malaria form vs critical form  ($2^{nd}$ case study): comparison of the 4 models, Tree, Prune,  $\lambda_{min}$-$L_1LR$ and $\lambda_{1se}$-$L_1LR$, through the three leave-one-out performance scores: recall, specificity and accuracy. For each score-bar, the color refers to the model, the length indicates the score's value (rounded percent) with its significance level for the recall and specificity scores.}
\label{fig:gtgperf4methods}
\end{figure}

Classification trees-based models (i.e. Tree and Prune) have a better accuracy than $L_1-LR$-based models (i.e. $\lambda_{1se}$-$L_1LR$ and  $\lambda_{min}$-$L_1LR$) for both case studies "cs 1 and 2" (see Fig. \ref{fig:stdperf4methods} and \ref{fig:gtgperf4methods}). The Tree method outperforms the other methods with an accuracy score of $68\%$ (resp. $70\%$)  to discriminate moderate and severe (resp. serious and critical) forms of  imported malaria. It is also the only method to have both highly significant recall and specificity scores (pvalue $\leq$ 0.001) for the two case studies. 
Indeed, $L_1-LR$-based models tend to well classify the second class of the case studies, that is the least represented one composed of the patients with severe (resp. critical) imported malaria in the first (resp. second) case study, displaying significant recall scores. On the other hand, they tend to fail to correctly classify the first class. 
Conversely, the Prune models well classify the first class of the case studies, the most represented class, composed of the patients with moderate (resp. serious) imported malaria in the first (resp. second) case study, having the best significant specificity scores. On the other hand, they fail to classify the second class correctly.
We can assume that the Tree method is more robust to unbalanced groups of samples and therefore extracts discriminant decision rules that generalize well to predict both classes. 

For both case studies, we also observed that the simpler forms of the standard methods, namely $\lambda_{1se}$-$L_1LR$ and Prune models, achieve the best significant recall and specificity scores, respectively. Conversely, they achieve the worst non-significant specificity and recall scores, respectively. Therefore, the sparsest approaches seem to be more sensitive to unbalanced groups of samples tending to over-classify a class more than the other.

\begin{figure}[h!]
\centering
\fbox{\includegraphics[width=3.5in]
{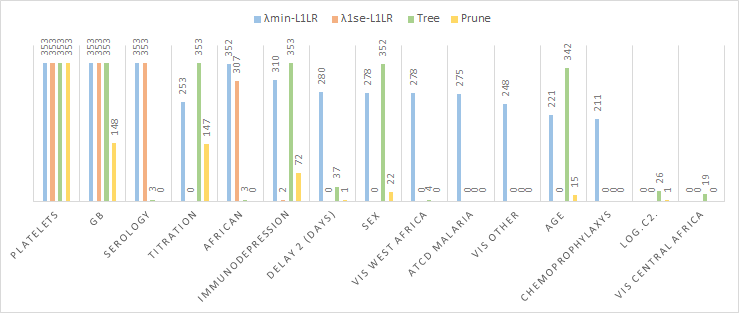}}
\caption{Moderate malaria form vs severe form ($1^{st}$ case study): frequency of the input variables selected by the four methods (Tree, Prune,  $\lambda_{min}$-$L_1LR$ and $\lambda_{1se}$-$L_1LR$), over the 353 models generated by the leave-one-out validation. We only represented features selected at least 10 times.}
\label{fig:4_methods_std}
\end{figure}

\begin{figure}[h!]
\centering

\fbox{\includegraphics[width=3.5in]
{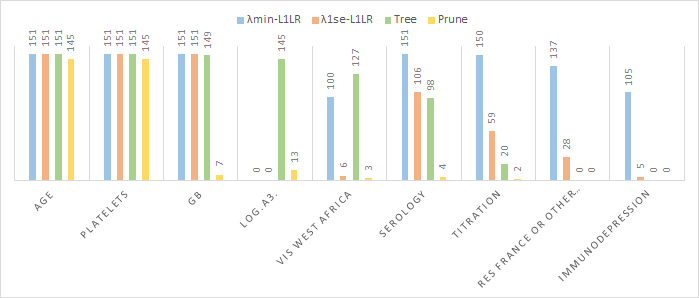}}
\caption{Serious malaria form vs critical form ($2^{nd}$ case study): frequency of the 
input variables selected by the four methods (Tree, Prune,  $\lambda_{min}$-$L_1LR$ and $\lambda_{1se}$-$L_1LR$), over the 353 models generated by the leave-one-out validation. We only represented features selected at least 10 times.}
\label{fig:4_methods_gtg}
\end{figure}

\paragraph{Selected variables}
As expected, the simpler models, namely $\lambda_{1se}$-$L_1LR$ and Prune, include less features than the standard models, namely $\lambda_{min}$-$L_1LR$ and Tree. For both the first case studies, and particularly the first one, the Tree-based models are on average sparser but less stable than the $L_1LR$-based models (see Fig. \ref{fig:4_methods_std} and \ref{fig:4_methods_gtg}). Indeed, they capture on average less features but some of them are selected only few times corresponding probably to locally optimal solutions.
A common pattern of selected stable features (i.e. almost selected systematically over the leave-one-out models) for all the methods is composed of white blood cells count (GB), platelets count, serological status and titration variables for "cs 1". We observed the same common pattern of selected stable features plus the age for "cs 2". Note that the serological status is a discrete feature deriving from the titration values and so they are considered similar features.

In the following, we focused on the Tree method, since it is the most powerful approach and it gives meaningful information on the models through the learned classification rules. Indeed, these latter characterize the different discriminant subregions of the feature space specific to subgroups of subjects.
In addition to the common pattern, the three models capture the following  stable features: the immunodependency and sex variables for "cs 1", and the log-transformed of the expression of  sub-group of \textit{var} gene family A and visit in West Africa variables for "cs 2". 

Some of these results confirmed the observations of previous studies on the potential interactions of \textit{Plasmodium falciparum} during acute malaria with negative hematological changes \citep{Muwonge:2013} like an increase of GB \citep{Berens-Riha:2014} and a decrease of platelets count \citep{Lampah:2014} and at the same time with an immunological protection represented by serological status \citep{Bouchaud:2005} on the development of the different severity forms of imported malaria. Furthermore, being older is a well-known risk factor for developing the acute form of imported malaria \citep{Checkley:2012}. In \citep{Checkley:2012}, a statistical relationship has been reported between visiting West Africa, especially Gambia, and the risk of fatal malaria.
Concerning the impact of gender on the discrimination between moderate and severe malaria, no statistical relation has been proven between gender and malaria severity. Nevertheless, one study showed that women are more susceptible to cerebral complications than men \citep{Muhlberger:2003}. Concerning the expression of group A \textit{var} gene, some studies have highlighted the role of the \emph{var} gene family in cerebral malaria \citep{Argy:2014}.

\subsubsection{Combined $L_1LR$-tree-based models}

\begin{figure}[h!]
\centering
\fbox{\includegraphics[width=3.5in]
{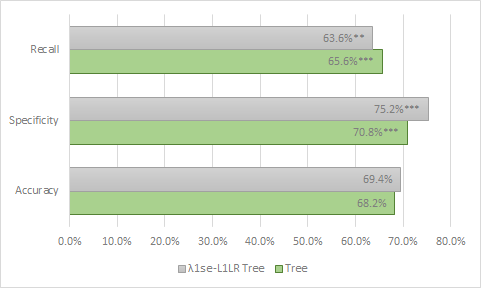}}
\caption{Moderate malaria form vs severe form ($1^{st}$ case study): comparison of the Tree and $\lambda_{1se}$-$L_1LR$ Tree models through the three leave-one-out performance scores: recall, specificity and accuracy. For each score-bar, the color refers to the model, the length indicates the score's value (rounded percent) with its significance level for the recall and specificity scores.}
\label{fig:stdperf3methods}
\end{figure}

\begin{figure}[h!]
\centering
\fbox{\includegraphics[width=3.5in]
{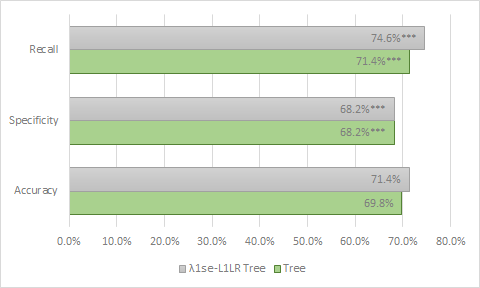}}
\caption{Serious malaria form vs critical form ($2^{nd}$ case study): comparison of the Tree and $\lambda_{1se}$-$L_1LR$ Tree models, through the three leave-one-out performance scores: recall, specificity and accuracy. For each score-bar, the color refers to the model, the length indicates the score's value (rounded percent) with its significance level for the recall and specificity scores.}
\label{fig:gtgperf3methods}
\end{figure}

To effectively penalize the Tree method with a prior $L_1LR$-based feature selection step, we used the $\lambda_{1se}-L_1LR$ method.   
\paragraph{Performance scores}
The combined $\lambda_{1se}-L_1LR$-tree method achieves similar or higher performances than the Tree ones, except for the recall score of the first case study which is $2\%$ inferior (see Fig. \ref{fig:stdperf3methods} and \ref{fig:gtgperf3methods}).

\begin{figure}[h!]
\centering
\fbox{\includegraphics[width=3.5in]
{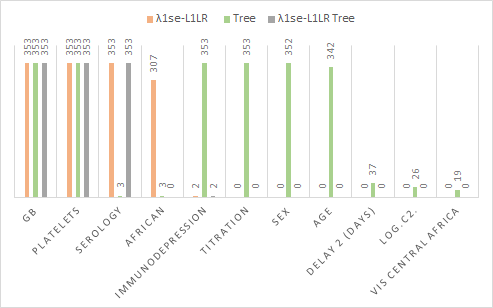}}
\caption{Moderate malaria form vs severe form  ($1^{st}$ case study): frequency of the input variables selected by the three methods ($\lambda_{1se}$-$L_1LR$, Tree and $\lambda_{1se}$-$L_1LR$ Tree), over the 353 models generated by the leave-one-out validation. We only represented features selected at least 10 times.}
\label{fig:stdselvar3}
\end{figure}

\begin{figure}[h!]
\centering

\fbox{\includegraphics[width=3.5in]
{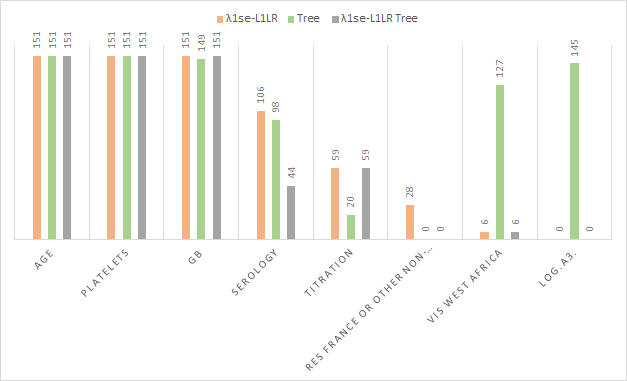}}
\caption{Serious malaria form vs critical form  ($2^{nd}$ case study): frequency of the
input variables selected by the three methods ($\lambda_{1se}$-$L_1LR$, Tree and $\lambda_{1se}$-$L_1LR$ Tree), over the 151 models generated by the leave-one-out validation. We only represented features selected at least 10 times.}
\label{fig:gtgselvar3}
\end{figure}

\paragraph{Selected variables}
As the combined method builds the classification tree based on the features selected with $\lambda_{1se}$-$L_1LR$, it efficiently reduces the set of input features. The set of stable variables selected by the combined models corresponds to the previously observed common patterns for both case studies: GB, platelets count and serological status/titration (resp. plus age) variables for "cs 1" (resp. "cs 2") (see Fig. \ref{fig:stdselvar3} and \ref{fig:gtgselvar3}). Note that for the second case study the 151 combined models have selected either serology or titration leading to a total frequency of 103 for both variables.
Therefore, the combined method led to sparser, more stable and discriminant (in terms of accuracy performances) models than those achieved by the Tree method.
Tables \ref{tab:1} and \ref{tab:2} show examples of rule sets derived from stable  $\lambda_{1se}$-$L_1LR$ Tree models for each case study. From these classification rules, we can easily point out the subregions of the feature space predictive of the severe forms of imported malaria.

\subsection{Second experiment}
Given the results of the methodological comparative study obtained on the two case studies of the first experiment, we applied the $\lambda_{1se}-L_{1}LR$ Tree to the four case studies of the second experiment.

\begin{figure}[h!]
\centering

\fbox{\includegraphics[width=3.5in]
{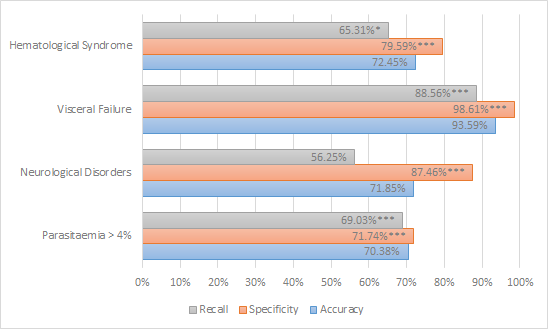}}
\caption{$2^{nd}$ experiment: leave-one-out performance scores (recall, specificity and accuracy) for the four clinical outputs: hematological syndrome, visceral failure, neurological disorders and parasitaemia level (cs 3 to 6). For each score-bar, the color refers to the performance score, the length indicates the score's value (rounded percent) with its significance level for the recall and specificity scores. The performance scores are grouped for clinical output.}
\label{fig:perf_clinical}
\end{figure}

\paragraph{Performance scores}
For all the case studies, the combined method discriminates with good accuracy scores between the two clinical states of the four clinical severity criteria (figure \ref{fig:perf_clinical}): hematological syndrome ($72.45\%$), visceral failure ($93.59\%$), neurological disorders ($71.85\%$) and parasitaemia level ($70.38\%$). 
We can also conclude that for all these case studies,  we significantly classify the two classes, except for the recall score of "cs 5" which can be explained by the low frequency of the class "Neurological disorders" (i.e. $9\%$).

\paragraph{Selected variables}

\begin{figure}[h!]
\centering

\fbox{\includegraphics[width=5.5in]
{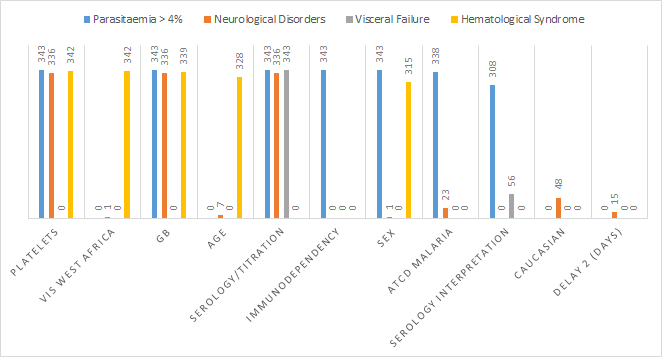}}
\caption{$2^{nd}$ experiment: frequency of the
input variables selected by the  $\lambda_{1se}$-$L_1LR$ Tree for the four clinical outputs: hematological syndrome, visceral failure, neurological disorders and parasitaemia level (cs 3 to 6), over the 343 models generated by the leave-one-out validation. We only represented features selected at least 10 times.}
\label{fig:var_clinical}
\end{figure}

The selected variables presented on Figure \ref{fig:var_clinical} and the classification rules (see Tables \ref{tab:3} to \ref{tab:6}) give medical insights on the influence of unused factors (demographic, epidemiological, clinical, biological and transcriptomic) on some clinical observations of acute malaria attacks. 
As previously observed in the first experiment, the variables platelets count and white blood cells count are strongly involved in the prediction of neurological disorders, hyper-parasitaemia and hematological syndrome. This could reflect the parasite sequestration in \textit{Plasmodium falciparum} malaria. Concerning "cs 5", the models showed that caucasian patients seem to be more affected by neurological disorders. Moreover, the corresponding classification rule set pointed out an interesting insight, that is the patients probably not previously affected by malaria (Caucasian, low titration/negative serology) are more sensitive to Neuro-malaria which indicates the presence of more severe forms of malaria \citep{Gupta:1999}. On the other hand, patients with a history of malaria indicated by a positive serology display visceral failures. We currently observed more frequently the moderate malaria form with visceral failures and without neurological disorders. Furthermore, the $\lambda_{1se}-L_{1}LR$ trees models of "cs 4" captured only the variable Serology as a predictive factor of visceral failures. This can be explained by the fact that these symptoms may arise more from an inflammatory or immunological response than from a parasite sequestration. The gender seems also to have an impact on the presence of the hematological syndrome and the hyper-parasitaemia. Indeed, the rule sets (see Tables \ref{tab:3} and \ref{tab:6}) show that for a given range of platelets count and a given GB threshold, male patients develop these clinical symptoms while women do not. This may be due to the fact that men travel more frequently than women in endemic areas.

\section{Conclusion and discussion}
\label{sec:Concl}
Among the standard approaches, i.e. $L_1$ logistic regression and regression trees, only the Tree 
method efficiently well classifies the two classes of patients for both the first and the second case studies. However, the Tree models are not sparse and stable enough providing locally optimal solutions reflecting the intrinsic heterogeneity of the studied dataset.
The pruning method drastically simplifies the Tree models while leading to poor, non-significant recall scores. This phenomenon could be explained by the fact that pruning tends to eliminate unstable branches, corresponding to variables with a great variance on threshold values and positions across cross-validation trees.  
Therefore, a pre-selection of the input features can be a good alternative solution to pruning in order to constrain the complexity and to increase the robustness to small data variations of the decision trees by removing under-represented phenomena in the studied population.

Our new method, called $\lambda_{1se}L_1LR$-tree, significantly
discriminates the two classes for both experiments and we show that it outperforms all the other methods in terms of accuracy for the two first case studies. Moreover, it efficiently leads to sparser and more stable models than the Tree ones. 
We can conclude that our combined method is a relevant sparse tree-based method for classification problems even when the classes are strongly unbalanced as it is the case for the classes of the second experiment. 

Concerning the prediction of the severe criteria of imported malaria, the combined method classifies around $70\%$ of the patients (until $93.59\%$ for the visceral failures) for both studied experiments. Hence, concerning the case study 2, we can conclude that the subclassification of severe imported malaria in serious and critical classes is valid. Moreover, the combined method produces explanatory and easily understandable models which can be represented under the form of rule sets. These rule sets confirm the predictive power of epidemiological and biological variables discarded from the current classification, such as platelets count, age, gender, white blood cells count and serology. They also provide meaningful information about the discriminant subregions of the selected features specifying for example the threshold or range of values of the selected biological measures.
  
However, these models did not capture some local phenomena in a stable way (cf variables captured with a low frequency over the leave-one-out models), probably due to their low representation in the dataset. This may explain a part of the misclassification of patients. A solution would be to expand the sample size, while ensuring the diversity of the population surveyed, in order to increase the statistical reliability of these phenomena. For the first experiment, a part of the classification error may also result from a bias in the definition of the classes based on the current clinico-biological picture. Indeed, as explained in the introduction, the diagnosis of severe imported malaria is multi-criteria, complex and does not take into account the heterogeneity of the individual profiles.

It is also important to mention that the use of the $\lambda_{1se}L_1LR$ as a feature selection step prior to fitting the decision tree may be challenged to overcome the limitations of the $L_1LR$ method (linear interactions, no missing data, etc.). In future work, it would be interesting to investigate other $L_1$ penalized approaches.

\clearpage

\section*{Appendix}

\begin{table}[h!]

\caption{moderate vs severe form ($1^{st}$ case study): classification rules of the two classes obtained with the $\lambda_{1se}-L_{1}LR$ tree model for a probability threshold equal to $0.4$.\label{tab:1}}
\begin{tabularx}{\textwidth}{ |X|X| }

\hline
   \multicolumn{2}{|c|}{\textbf{First Case Study: Classification Rules}} \\
\hline
\centering\textbf{Moderate Form} & \centering{\textbf{Severe Form}} \\
\tabularnewline
\hline 

$Platelets\geq84,\,GB\leq7 $& $Platelets\geq84,\,GB\in[7;7.9],\,Serology=Positive$\\
\hline
$Platelets\geq84,\,GB\geq7.9,\,Serology=Positive $& $Platelets\geq84,\,GB\geq 7,\,Serology=Negative$\\
\hline
$Platelets\in[46;84],\,GB\leq 4.9,\,Serology=Positive $& $Platelets\in[46;84],\,Serology=Negative$\\
\hline
$Platelets\in[46;68],\,GB\geq6.1,\,Serology=Positive $& $Platelets\in[68;84],\,GB\geq 4.9,\,Serology=Positive$\\
\hline
& $Platelets\in[46;68],\,GB\in [4.9;6.1],\,Serology=Positive$\\
\hline
& $Platelets<46$\\
\hline
\end{tabularx}
\end{table}

\begin{table}[h!]

\caption{Serious vs critical form ($2^{nd}$ case study): classification rules of the two classes obtained with the $\lambda_{1se}-L_{1}LR$ tree model for a probability threshold of $0.4$.\label{tab:2}}
\begin{tabularx}{\textwidth}{ |X|X| }
\hline
   \multicolumn{2}{|c|}{\textbf{Second Case Study: Classification Rules}} \\
\hline
\centering\textbf{Serious Form} & \centering{\textbf{Critical Form}} \\
\tabularnewline
\hline 

$Age\in [16;44],\,Platelets\geq30,\,Serology=Negative $& $Age<16,\,Platelets\geq30$\\
\hline
$Age\in [16;44],\,Platelets\geq30,\,Serology=Positive,\,GB<5.9 $& $Age\in[16;44],\,Platelets\geq30,\,Serology=Positive,\,GB\geq 5.9$\\
\hline
$Age\geq50,\,Platelets\geq52 $& $Age<44,\,Platelets<30$\\

\hline
& $Age \in[44;50],\,Platelets\geq 52$\\
\hline
& $Age\geq44,\, Platelets<52$\\
\hline
\end{tabularx}
\end{table}

\begin{table}[h!]

\caption{No Hematological Syndrome vs Hematological Syndrome ($3^{rd}$ case study): classification rules of the two classes obtained with the $\lambda_{1se}-L_{1}LR$ tree model for a probability threshold of $0.15$.\label{tab:3}}
\begin{tabularx}{\textwidth}{ |X|X| }
\hline
   \multicolumn{2}{|c|}{\textbf{Third Case Study: Classification Rules}} \\
\hline
\centering\textbf{No Hematological Syndrome} & \centering{\textbf{Hematological Syndrome}} \\
\tabularnewline
\hline 

$Platelets\geq 58$& $Platelets<40$\\
\hline
$Platelets\in [40;58],\,GB\geq3.1,\,Sex=Female $& $Platelets\in [40;58],\,GB\geq3.1,\,Sex=Male $\\
\hline
& $Platelets\in [40;58],\,GB<3.1$\\

\hline
\end{tabularx}

\end{table}

\begin{table}[h!]

\caption{No Visceral Failure vs Visceral Failure ($4^{th}$ case study): classification rules of the two classes obtained with the $\lambda_{1se}-L_{1}LR$ tree model for a probability threshold of  $0.79$.\label{tab:4}}

\begin{tabularx}{\textwidth}{ |X|X| }
\hline
   \multicolumn{2}{|c|}{\textbf{Fourth Case Study: Classification Rules}} \\
\hline
\centering\textbf{No Visceral Failure} & \centering{\textbf{Visceral Failure}} \\
\tabularnewline
\hline 
$Serology = Negative$& $Serology = Positive$\\
\hline
\end{tabularx}
\end{table}

\begin{table}[h!]

\caption{No Neurological Disorders vs Neurological Disorders ($5^{th}$ case study): classification rules of the two classes obtained with the $\lambda_{1se}-L_{1}LR$ tree model for a probability threshold of $0.09$.\label{tab:5}}
\begin{tabularx}{\textwidth}{ |X|X| }

\hline
   \multicolumn{2}{|c|}{\textbf{Fifth Case Study: Classification Rules}} \\
\hline
\centering\textbf{No Neurological Disorders} & \centering{\textbf{Neurological Disorders}} \\
\tabularnewline
\hline 
$Platelets\in[26;46], \, GB<6.8,\, Caucasian = False$& $Platelets\in[26;46], \, GB<6.8,\, Caucasian = True$\\
\hline
$Platelets\geq 46,\,Titration>64 $& $Platelets<46,\,GB\geq6.8 $\\
\hline
$Platelets\geq 46,\,Titration\leq64,\, GB\in [4;7.2] $& $Platelets\geq 46,\,Titration\leq64,\, GB<4  $\\
\hline
& $Platelets< 26,\, GB<6.8  $\\
\hline
& $Platelets\geq 46,\, Titration\leq 64,\, GB\geq7.2  $\\
\hline
\end{tabularx}
\end{table}

\begin{table}[h!]

\caption{Parasitology $\leq$ 4\% vs Parasitology \textgreater 4\% ($6^{th}$ case study): classification rules of the two classes obtained with the $\lambda_{1se}-L_{1}LR$ tree model for a probability threshold equal to $0.33$.\label{tab:6}}
\begin{tabularx}{\textwidth}{ |X|X| }
\hline
   \multicolumn{2}{|c|}{\textbf{Sixth Case Study: Classification Rules}} \\
\hline
\centering\textbf{Parasitology $\leq$ 4\% } & \centering{\textbf{Parasitology \textgreater 4\%}} \\
\tabularnewline
\hline 
$Platelets\in[20;40], \, GB\in[5.5;6.8]$& $Platelets<20$\\
\hline
$Platelets\geq 40,\,GB>6.9,\, ATCD = True $& $Platelets\in[20;40],\,GB\leq5.5 $\\
\hline
$Platelets\geq 40,\,GB>8,\, ATCD = False,\, Serology = Positive$& $Platelets\geq 40,\,GB>8,\, ATCD = False,\, Serology = Negative$\\
\hline
$Platelets\geq 84,\, GB<6.9,\, Immunodependency = False$& $Platelets\geq 84,\, GB<6.9,\, Immunodependency = True  $\\
\hline
$Platelets\in [40;84],\, GB<6.9,\, Sex = Female$& $Platelets\in [40;84],\, GB<6.9,\, Sex = Male$\\
\hline
& $Platelets\geq 40,\, GB\in [6.9;8],\, ATCD = False,\, Serology = Positive $\\
\hline
& $Platelets\in [20;40],\, GB\geq 6.8$\\
\hline
\end{tabularx}
\end{table}

\end{document}